\documentclass[10pt,twocolumn,letterpaper]{article}

\usepackage{iccv,times,graphicx,tabulary,multirow,subfig,xspace,xcolor}
\usepackage{amsthm,amssymb,fixmath,mathtools,nicefrac}
\usepackage[pagebackref=true,breaklinks=true,letterpaper=true,colorlinks,
  citecolor=citecolor,bookmarks=false]{hyperref}

\definecolor{citecolor}{RGB}{34,139,34}

\newcommand{\todo}[1]{}
\renewcommand{\todo}[1]{{\color{red}{#1}}}

\newcommand{\comment}[1]{}
\renewcommand{\comment}[1]{{\color{blue}comment:{#1}}}

\newcommand{\nanzhu}[1]{}
\renewcommand{\nanzhu}[1]{{\color{orange}{#1}}}

\newcommand{\yogesh}[1]{}
\renewcommand{\yogesh}[1]{{\color{green}{#1}}}

\newcommand{\mehmet}[1]{}
\renewcommand{\mehmet}[1]{{\color{purple}{#1}}}

\newcommand{\jan}[1]{}
\renewcommand{\jan}[1]{{\color{cyan}{#1}}}

\newcommand{\dataset}[1]{\emph{#1}}

\begin{document}
\title{Ensembling Instance and Semantic for Panoptic Segmentation}
\track{COCO Panoptic Segmentation}

\author{Mehmet Yildirim, Yogesh Langhe, Jan Richter, Nanzhu Jiang \\
Yaroslav Tarasenko, Lyubov Klein, Claus Bahlmann\\
Siemens Mobility\\
{\tt\small \{mehmet.yildirim, claus.bahlmann\}@siemens.com}
\and
}

\maketitle


\begin{abstract}

In this technical report, we demonstrate our solution for the 2019 COCO panoptic segmentation task. Our method first performs instance segmentation and semantic segmentation separately, then combines the two to generate panoptic segmentation results. To enhance the performance, we add several expert models of Mask R-CNN in instance segmentation to tackle the data imbalance problem in the training data; also HTC model is adopted yielding our best instance segmentation results.
In semantic segmentation, we trained several models with various backbones and use an ensemble strategy which further boosts the segmentation results. In the end, we analyze various combinations of instance and semantic segmentation, and report on their performance for the final panoptic segmentation results.
Our best model achieves $PQ$ 47.1 on 2019 COCO panoptic test-dev data.

\end{abstract}


%


\section{Introduction}

Panoptic segmentation is a challenging task in the computer vision community which can help solving problems of scene understanding including autonomous driving. The coherent scene segmentation problem can be solved by combining instance and semantic segmentation. The aim for instance segmentation is to first detect and locate all foreground objects by bounding boxes, then for a detected object, assign a category label and an index number on all related pixels of that object. The goal of semantic segmentation is to classify all pixels in an image into different category labels, whereas it does not consider indices of objects. Both tasks have their own limitations for image segmentation. In order to generate a dense and coherent scene segmentation, panoptic segmentation was introduced in \cite{kirillov2019panoptic} combining the strengths of instance and semantic segmentation with the aim to assign a category label and an instance index for each pixel of a given image.

We use MS COCO\cite{lin2014coco} dataset for this challenge. For image segmentation tasks, MS COCO provides both training and validation data in the form of images and comprehensive annotations for all three segmentation tasks. These data are commonly used by the vision community, and related vision competitions for object detection and segmentation are yearly held.


In our solution to panoptic segmentation, we use a standard approach of combining instance and semantic segmentation as described in \cite{kirillov2019panoptic}.
To realize this, we adopted the commonly used Mask R-CNN\cite{DBLP:conf/iccv/HeGDG17} and the powerful
Hybrid Task Cascade (HTC)\cite{DBLP:journals/corr/abs-1901-07518, mmdetection} for instance segmentation.
Mask R-CNN extends Faster R-CNN\cite{DBLP:journals/corr/RenHG015} originally designed for object detection by adding a separate branch for mask prediction, thereby predicting pixel level object masks in parallel to predicting the bounding box positions of objects.
The HTC framework offers a special cascaded combination of R-CNN and gives the best performance in 2018 COCO object detection challenge\footnote{\url{http://cocodataset.org/##detection-leaderboard}, in SEGM Chal18}.
%
%
In semantic segmentation, we adopted DeepLabv3\cite{DBLP:journals/corr/ChenPSA17} network, which combines concepts of spatial pyramid pooling and dilated convolutions to extract information in the image at different scales and to capture information from a larger effective field of view. 
%
In the final combination step, we use the panopticAPI\cite{kirillov2019panoptic} to combine the predictions generated from instance and semantic models to yield the final panoptic results.

For our implementation, we adopted the open source machine learning framework Pytorch\cite{paszke2017automatic} as well as the accompanying torchvision library. They are widely used by the vision community, which allows us to quickly experiment with different pre-trained models and network architectures.

Segmentation experiments were conducted on a fully automated cloud pipeline on AWS including training, prediction, evaluation, as well as real time visualization of segmentation results.



\section{Method Description}
\subsection{Dataset}

In MS COCO dataset, the training data \dataset{train2017}(115k images) along with annotations for instance, semantic and panoptic segmentation respectively. The validation data \dataset{val2017} consists of $5k$ images, also with annotations. 

All our experiments are trained on \dataset{train2017}, and the reported instance and semantic evaluation results are validated on \dataset{val2017}. For panoptic segmentation, we also report on \dataset{test-dev}, where we get the results from codalab submission page\footnote{https://competitions.codalab.org/competitions/19507}.
No external data is used.

Note that categories belong to stuff\cite{caesar2018coco} in the COCO stuff annotation are not the same as those in COCO panoptic annotations.
We therefore use the converted stuff annotations from panoptic annotations instead of the original stuff annotations. We also treated all categories belonging to ``thing" as one ``merged-thing" category.

\subsection{Baseline}

In the training session of our baseline method for panoptic segmentation, we trained a single model for instance segmentation using Mask R-CNN\cite{DBLP:conf/iccv/HeGDG17}, and a single model for semantic segmentation using DeepLabv3\cite{DBLP:journals/corr/ChenPSA17} architecture. %
For feature extraction, after examine the performance of several backbones, we select the ResNet152\cite{DBLP:journals/corr/HeZRS15} as the backbone for Mask R-CNN and ResNet101 for DeepLabv3.
No ensemble, test time tricks or post processing tricks at any level used in this baseline.

Note that for all experiments including the baseline, the optimizer is consistent. We use Stochastic Gradient Descent(SGD) with momentum. For instance segmentation, a multi-step learning scheduler (LR) with initial warm-up is adopted, whereas for semantic segmentation, an additional LR is used to generate a decaying LR curve using a multiplicative factor.

Our panoptic segmentation builds upon instance and semantic segmentation therefore the performance of these two has crucial influence on the final panoptic segmentation results. In the following section, we will report in detail, how we improve this baseline by studying different strategies we used for instance and semantic segmentation respectively.

\subsection{Instance Segmentation}

\subsubsection{Mask R-CNN}

To increase the instance segmentation accuracy, we performed the following strategies upon our baseline using Mask R-CNN.

\textbf{Backbone.} In order to select the backbone that performed best with Mask R-CNN\cite{DBLP:conf/iccv/HeGDG17}, we examined the performance of the commonly used backbone ResNet50 and ResNet152 with Mask R-CNN. As can be seen in Table \ref{table:results_instance},
the best performance is achieved with ResNet152, therefore it was selected for multi-scale feature extraction.

\textbf{Expert Models.} The training data of instance segmentation illustrates highly imbalanced distribution for the categories. In particular, the most annotated categories are ``person" and ``car" that may lead to inaccurate category bias of training models\cite{DBLP:journals/corr/abs-1804-10851}. To conquer this problem, we divided the \dataset{train2017} datasets in three subsets respectively, with the first containing only annotations for ``person" category, the second for ``car", and the third for all other categories. We then trained three expert models using Mask R-CNN framework with ResNet152 backbone on each of the subsets. By doing so, we achieved an increase of $0.8$ mAP as shown in Table \ref{table:results_instance}.

\subsubsection{Hybrid Task Cascade Framework}

Besides of the Mask R-CNN framework, we also tested the Hybrid Task Cascade(HTC) framework\cite{DBLP:journals/corr/abs-1901-07518, mmdetection} for instance segmentation. 
We use HTC with X-101-64x4d-FPN backbone and deformable convolution in the last stage (res5) of the backbone. 
Using this model, we get the highest bbox mAP for instance segmentation, as shown in Table \ref{table:results_instance}

\subsection{Semantic Segmentation}

In our semantic segmentation, we tested FCN\cite{DBLP:journals/corr/LongSD14} and DeepLabv3\cite{DBLP:journals/corr/ChenPSA17}. As shown in Table \ref{table:results_semantic}, the segmentation accuracy of FCN provides 6.7 mIoU lower compared to DeepLabv3 baseline, therefore we choose DeepLabv3 as our network. 
In order to examine its performance, we trained multiple DeepLabv3 models with different backbones such as ResNet101 and ResNet152. Note that in  the training session, we combine all thing categories as one category for ``merged-thing".

We find that the performance of all single models converged similarly at around $43$ mIoU. For example, with ResNet101 we get mIoU of $43.4$, with ResNet152 we get mIoU of $43.2$. In order to improve the accuracy, we ensemble multiple models by taking the average of their pixel-wise confidence map for each category. We use our top performing semantic models for ensembling to improve performance. The results will be discussed in the next section.





\begin{table}[t]
\footnotesize
\centering
\begin{tabular}{|c| c| c| c|} 
\hline
Strategy & Instance Segmentation & mAP (bbox) & mAP (mask) \\
\hline
Ins1 & MaskRCNN ResNet50 & 37.9 & 34.6 \\ 
Ins2 & MaskRCNN ResNet152 & 41.3 & 37.1 \\
Ins3 & expert models for Ins2 & 42.1 & 38.0 \\
\hline
Ins4 & HTC + X-101-64x4d-FPN & 50.7 & 43.9 \\
\hline
\end{tabular}
\caption{Instance segmentation results for various strategies tested on \dataset{val2017}.}
\label{table:results_instance}
\end{table}


\begin{table}[t]
\footnotesize
\centering
\begin{tabular}{|c| c| c|} 
\hline
Strategy & Semantic Segmentation & mIoU \\
\hline
Sem1 & FCN ResNet50 & 36.7 \\
Sem2 & DeepLabv3 ResNet101 & 43.4 \\
Sem3 & DeepLabv3 ResNet152 & 43.2 \\ 
Sem4 & Ensemble multiple models & 44.5 \\
\hline
\end{tabular}
\caption{Semantic segmentation results for various strategies tested on \dataset{val2017}.}
\label{table:results_semantic}
\end{table}


\begin{table*}[t]
\small
\centering
\begin{tabular}{|c| c| c| c| c|} 
\hline
Strategy & Instance & Semantic & PQ (\dataset{val2017})& PQ (\dataset{test-dev}) \\
\hline
Baseline & MaskRCNN ResNet152 & DeepLabv3 ResNet101 & 42.7 & 42.9 \\
\hline
Pan1 & + Expert Models & - & 43.7 & 43.7 \\ 
\hline
Pan2 & - & + Ensemble 3 Models & 43.2 & 43.4 \\
\hline
Pan3 & + Expert Models & + Ensemble 3 Models & 44.2 & 44.2 \\
\hline
Pan4 & HTC & + Ensemble 3 Models & 46.7 & 47.1 \\
\hline
\end{tabular}
\footnotesize
\caption{Panoptic evaluation results for various experiments tested on \dataset{val2017} and \dataset{test-dev}.}
\label{table:results_panoptic}
\end{table*}

\section{Experiments and Results}


In this section, we report our experiment settings and evaluation results.
%
In the first step, we conducted experiments on instance and semantic segmentation using the standard evaluation measure of mean average precision (mAP) and mean Intersection over Union (mIoU), respectively. In the second step, we tested the various combination of them to study how we achieve our best panoptic segmentation results. All the following experiments are trained on dataset \dataset{train2017}, and the reported evaluation results are validated on \dataset{val2017}. No external data is used.
All trainings were executed on 8xGPU V100 AWS EC2-instances and testing is done locally on GPU machines with 4x Titan Xp.

We represent various experimental setup in Table \ref{table:results_instance}. Our first setup denoted as Ins1 is the trained Mask R-CNN with ResNet50 backbone, where we get mAP $37.9$ for bounding box (bbox), and mAP $34.6$ for mask.
%
Then, we changed the backbone to ResNet152 in Ins2 and get mAP for bbox $41.3$ and for mask $37.1$. This shows ResNet152 outperforms ResNet50 for Mask R-CNN. Next, we tackle the data imbalance problem in Ins3, and trained three expert models using Mask R-CNN with ResNet152 on three subsets of training data. This yields a further increase in mAP of $0.8$ for bbox and $0.9$ for mask compared to Ins2, which shows that expert models indeed help for the case of data imbalance. Furthermore, with Ins4 strategy, we adopt HTC in our pipeline to improve instance performance. We achieve bbox $50.7$ mAP and mask $43.9$ mAP using X-101-64x4d-FPN backbone.

In the first semantic segmentation experimental setup (Sem1: Table \ref{table:results_semantic}), we get mIoU $36.7$ for FCN architecture, which is less than the other setups using DeepLabv3. In Sem2 and Sem3, we test DeepLabv3 with backbones ResNet101 and ResNet152 respectively. Their performance is quite similar, converging to mIoU around $43$. Although the model with the larger backbone (ResNet152) promises better mIoU score, we could not observe an improvement compared to ResNet101 in our benchmarks. Another approach is to ensemble three DeepLabv3 models in Ins4, one with ResNet101, the other two with ResNet152, yielding our highest mIoU of $44.5$. Compared to the best single semantic model, we improve mIoU of 1.1 using ensemble technique.


\begin{figure*}
  \centering
  \includegraphics[width=0.8\linewidth]{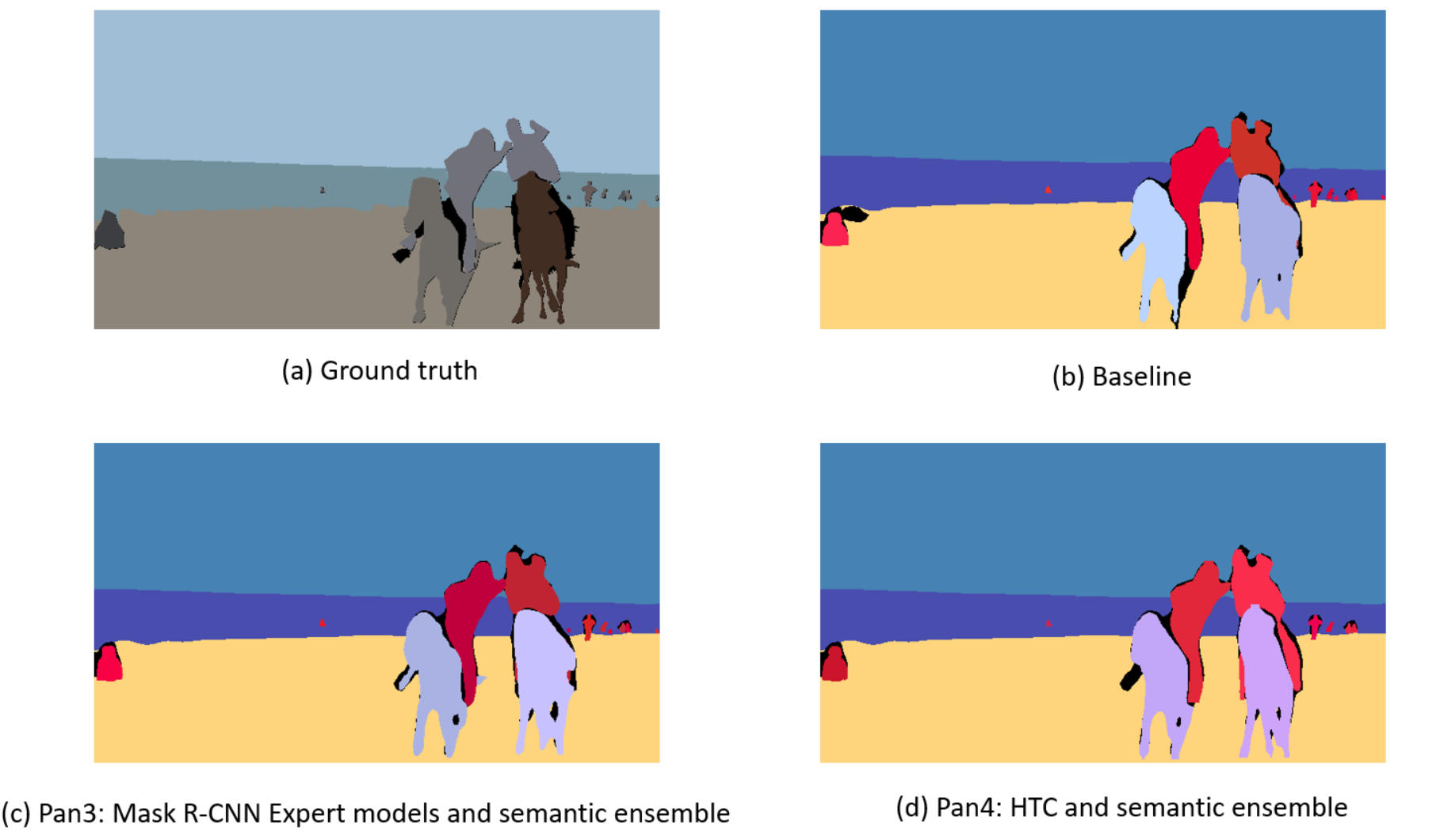}
  \caption{Illustration of different panoptic segmentation strategies. }
\label{fig:pan_figure}
\end{figure*}

We now discuss our final panoptic segmentation results by various combination of instance and semantic segmentation.
As shown in Table \ref{table:results_panoptic},
our chosen baseline using the best single model of Mask R-CNN and DeepLabv3 yields PQ of $42.7$ on dataset \dataset{val2017} and
$42.9$ on dataset \dataset{test-dev}. 
In setup Pan1 we enhanced the baseline by adding expert models in instance segmentation yielding a higher PQ in $43.7$ for both data-sets. Similarly, in setup Pan2 we enhanced the semantic segmentation using ensemble models, but keep instance segmentation unchanged as the baseline. 
We get $43.2$ and $43.4$ for the two data-sets respectively. This confirms that expert models and ensemble strategy enhanced the baselines in different aspects. Next, we use both enhancements in strategy Pan2 and Pan3 to achieve our highest result for Mask R-CNN and DeepLabv3 with PQ of $44.2$ on both data-sets. 
Finally, changing the instance framework to HTC, we get our highest PQ for the panoptic segmentation challenge for $46.2$ and $47.1$ on the two data-sets, respectively. This means that HTC not only contributes in instance segmentation but also have a crucial influence on panoptic results.

As an image example to illustrate our panoptic segmentation result, we show in Figure \ref{fig:pan_figure} for a comparison between baseline method and enhanced methods. Their segmentation behavior among different categories are roughly similar. However, strategy Pan4 works better than other methods by detecting small parts of object like persons' leg as shown in Figure \ref{fig:pan_figure}d.

\section{Conclusion}

We achieved PQ $47.1$ for the 2019 COCO panoptic segmentation task using combined approach of instance and semantic segmentation. In our study, we proofed that ensemble techniques and expert models indeed improve panoptic segmentation quality. Albeit our combined segmentation approach is computationally expensive compared to unified approach, it is more flexible in incorporating and improving on recent developments in both instance and semantic segmentation individually. Our future work may include more investigation on expert models, ensemble techniques and more framework or architecture, where we believe that there is still space for improvement in these aspects.

  
{\small \bibliographystyle{ieee_fullname} \bibliography{egbib}}

\end{document}